\documentclass[screen,authorversion]{acmart}

\usepackage[linesnumbered,ruled,vlined]{algorithm2e}

\usepackage{float}

\copyrightyear{2025}
\acmYear{2025}
\setcopyright{rightsretained}
\acmConference[LAK 2025]{LAK25: The 15th International Learning Analytics and Knowledge Conference}{March 03--07, 2025}{Dublin, Ireland}
\acmBooktitle{LAK25: The 15th International Learning Analytics and Knowledge Conference (LAK 2025), March 03--07, 2025, Dublin, Ireland}
\acmPrice{}
\acmDOI{10.1145/3706468.3706498}
\acmISBN{979-8-4007-0701-8/25/03}

\begin{document}

\title[ABROCA Distributions For Algorithmic Bias Assessment]{ABROCA Distributions For Algorithmic Bias Assessment: Considerations Around Interpretation}

\author{Conrad Borchers}
\affiliation{
  \institution{Carnegie Mellon University}
  \streetaddress{5000 Forbes Ave}
  \city{Pittsburgh, PA 15213}
  \country{USA}
}
\email{cborcher@cs.cmu.edu}

\author{Ryan S. Baker}
\affiliation{
  \institution{University of Pennsylvania}
  \streetaddress{3101 Walnut St}
  \city{Philadelphia, PA 19104}
  \country{USA}
}
\email{ryanshaunbaker@gmail.com}

\renewcommand{\shortauthors}{Borchers and Baker}

\begin{abstract}
Algorithmic bias continues to be a key concern of learning analytics. We study the statistical properties of the Absolute Between-ROC Area (ABROCA) metric. This fairness measure quantifies group-level differences in classifier performance through the absolute difference in ROC curves. ABROCA is particularly useful for detecting nuanced performance differences even when overall Area Under the ROC Curve (AUC) values are similar. We sample ABROCA under various conditions, including varying AUC differences and class distributions. We find that ABROCA distributions exhibit high skewness dependent on sample sizes, AUC differences, and class imbalance. When assessing whether a classifier is biased, this skewness inflates ABROCA values by chance, even when data is drawn (by simulation) from populations with equivalent ROC curves. These findings suggest that ABROCA requires careful interpretation given its distributional properties, especially when used to assess the degree of bias and when classes are imbalanced.
\end{abstract}



\maketitle

\section{Introduction}

Algorithmic bias across learner groups and contexts has been an increasing topic of interest in learning analytics, most notably because of the pervasiveness of predictive models in learning analytics and their high-stakes nature \cite{xu2024contexts}. Algorithmic bias is of critical concern within education as it could undermine the effectiveness of learning analytics. While different definitions and conceptualizations of algorithmic bias and fairness exist \cite{baker2022algorithmic}, their common denominator typically revolves around systematic unfairness or unequal treatment of groups caused by algorithms. This bias occurs when an algorithm produces results that disproportionately disadvantage or favor particular groups of people based on non-malleable characteristics like race, gender, or socioeconomic status \cite{friedman1996bias}. 

Recent learning analytics research argued that although the vast majority of published papers investigating algorithmic bias in education find evidence of bias \cite{baker2022algorithmic}, \textit{some} predictive models appear to achieve fairness, with minimal difference in model quality across demographic groups. For example, Zambrano et al. \cite{zambrano2024investigating} evaluated careless detectors and Bayesian knowledge tracing models, finding near-equal performance across groups defined by race, gender, socioeconomic status, special needs, and English language learner status. Similarly, Jiang and Pardos \cite{jiang2021towards} compared accuracies of grade prediction models across ethnic groups, concluding that an adversarial learning approach led to the fairest models but did not engage in the question of whether their fairest model was sufficiently fair. Indeed, one of the challenges to concluding that a model is unbiased is deciding which measures of bias to use. As noted by Kizilcec and Lee \cite{kizilcec2022algorithmic}, many possible ways exist to measure (or theoretically frame) algorithmic bias. 

ABROCA \cite{gardner2019evaluating}, or Absolute Between-ROC Area, has become popular for assessing algorithmic bias within our community. ABROCA captures the discrepancy between the model's performance across various subgroups. The metric highlights inconsistencies between models that may not be apparent when comparing only overall AUC values for different groups. While both ABROCA and AUC differences are meaningful, they highlight different aspects of a model's performance. A large ABROCA value alongside no difference in group AUC values suggests that the model behaves differently in specific parts of the data, even though it appears fair overall. This means that ABROCA can reveal subtle biases that simple comparisons of AUC might miss. However, it is possible that AUC ROC curves sometimes vary due to random chance, producing spuriously high ABROCA values. Thus it is important to understand whether an ABROCA value is large enough to indicate bias in certain regions of the model's performance. By understanding how ABROCA behaves when there is no bias, we can determine when an observed ABROCA is large enough to be meaningful. 

To understand when ABROCA differences are meaningful, we study ABROCA in relation to AUC ROC through simulation using sample size specifications typical for learning analytics. First, we study relationships between ABROCA and AUC differences where there is (by simulation) no or little true difference in classifier performance between two groups. This addresses the open question of how confident researchers can conclude whether a classifier is biased given different ABROCA values. Second, we systematically vary sample sizes and true differences in AUC values by group to examine when research can reliably identify group-level classifier performance differences using ABROCA. Third, we systematically vary the distribution of the outcome class and minority-class size. Many learning analysis prediction tasks have imbalanced classes (e.g., dropout prediction, with dropout being comparatively rare), and minority classes of interest (e.g., ethnic or racial minorities) are also commonly underrepresented in real-world data sets. It is plausible that spurious systematic differences in ABROCA may become more exaggerated under unbalanced classes (e.g., because estimating performance is harder with fewer data points). However, these possibilities have not yet, to the best of our knowledge, been documented. Therefore, we seek to answer the following research questions (RQs): 

\begin{itemize}
\item \textbf{RQ1:} What is the distribution of ABROCA under no difference in AUCs between two groups?
\item \textbf{RQ2:} How do ABROCA distributions vary as differences in AUC ROC get bigger?
\item \textbf{RQ3:} How do these ABROCA distributions vary by minority class and the outcome class imbalance?
\end{itemize}

We contribute open-source code to simulate ABROCA under different sample specifications.\footnote{https://github.com/conradborchers/abroca-sim} Further, we demonstrate that (a) ABROCA tends to be skewed toward higher values for smaller sample sizes, imbalanced outcome classes, and imbalanced minority student classes, making it more difficult to reliably determine if a classifier is algorithmically biased in those cases, (b) that ABROCA converges to the differences in AUC for larger samples, and (c) contribute to the understanding of how to interpret the ABROCA metric compared to differences in AUC ROC. This paper's goal is not to argue for ABROCA nor against ABROCA's merits compared to other approaches but to caution that ABROCA requires careful interpretation as an algorithmic bias metric given its distributional properties, especially when used to assess the degree of algorithmic bias in a classifier, rather than to identify potential areas of concern for an algorithm.

\section{Related Work}

\subsection{Algorithmic Bias and Fairness in Education}

Algorithmic bias remains a critical issue in learning analytics, where fairness in model performance across diverse student populations is essential to ensure equitable educational outcomes. Bias in machine learning systems is broadly defined as the systematic and unfair discrimination against specific individuals or groups in favor of others \cite{friedman1996bias}. There are three broad paradigms of understanding algorithmic bias in education: similarity-based, causal, and statistical \cite{verma2018fairness,kizilcec2022algorithmic,barocas2020hidden}. Similarity-based fairness emphasizes treating students with similar profiles equally, ensuring that predictions are consistent for individuals with comparable feature distributions. Causal fairness models the underlying causal relationships between sensitive attributes and predictions. Using counterfactual reasoning, causal fairness evaluates whether altering a student's sensitive attribute would change the prediction outcome \cite{verma2018fairness}.

Statistical fairness examines whether the predicted outcomes are independent of group membership, such as race or gender. Most recent research in learning analytics falls under this paradigm, most commonly under the lens of \textit{predictive performance-oriented fairness}. This form of fairness consists of setting a goal for the model's predictions to demonstrate equal performance metrics—such as accuracy, precision, or recall—across demographic groups (e.g., different races, genders, or socioeconomic statuses). This definition of statistical fairness is also known as separation, requiring an algorithm's decision to be independent of group membership conditional on true outcomes (for an overview, see Kizilcec \& Lee \cite{kizilcec2022algorithmic}). Recent work in EDM, LAK, and related communities attempted to create variants of these paradigms that look for broader ranges of differences, including ABROCA for separation \cite{gardner2019evaluating} and MADD for demographic parity \cite{verger2023your}.

To better understand ABROCA, we consider its most common use as a statistical separation metric in the statistical fairness paradigm to assess overall classifier fairness, comparing it to the most related metric in that paradigm, the difference in AUC ROC. We contribute to the literature by exploring how ABROCA behaves under different conditions and how ABROCA's properties warrant careful interpretation as a fairness metric in machine learning models for education. Through our simulations, we aim to increase understanding of ABROCA for use in research and practice. 

\subsection{Development and Use of Fairness Metrics in Learning Analytics}

The last decade has seen the development of a range of metrics for evaluating algorithmic bias and fairness across fields (see review in Kizilcec \& Lee \cite{kizilcec2022algorithmic}). The most common method for evaluating algorithmic bias and fairness in learning analytics has been to compare the same metric for different groups under the statistical fairness paradigm (see review in Baker \& Hawn \cite{baker2022algorithmic}). However, other metrics have seen considerable use. Equalized odds, also known as conditional procedure accuracy equality, is another widely used metric in learning analytics, focusing on balancing true positive and true negative rates between groups to ensure equal treatment across decision thresholds \cite{xu2024contexts,deho2024past}. Verger et al. \cite{verger2023your} proposed the Model Absolute Density Distance (MADD) to examine discriminatory behaviors in models independent of their predictive accuracy and the metric has been used in papers within our field \cite{vsvabensky2024evaluating}. Finally, Xu et al. \cite{xu2024contexts} propose another new fairness measure: using pseudo $R^2$ to predict the correctness of individual predictions based on demographic group attributes. In this case, less variation in model residuals explained by these attributes is desirable.

As noted above, ABROCA has become a popular metric in learning analytics. Gardner et al. \cite{gardner2019evaluating} propose ABROCA as a method for assessing performance differences between two classifiers along different precision-recall tradeoff thresholds that capture more nuance than only comparing AUC ROC values. ABROCA's calculation takes account of regions where a model may perform unequally across groups. Many learning analytics studies have used ABROCA; for example, Xu et al. \cite{xu2024contexts} used ABROCA to quantify fairness in predictive model transfer for higher education academic performance prediction. Notably, Xu et al. used a Wilcoxon signed-rank test to compare ABROCA values between two predictive models applied to the same data, which is relevant to our study of the reliability and randomness of ABROCA variations in data. Sha et al. \cite{sha2021assessing} used ABROCA to study gender differences in educational forum post-classification. Deho et al. \cite{deho2024past} used ABROCA to study the impact of dataset drift on the fairness of course grade prediction models. 

As the field of learning analytics continues to grow, it is crucial to understand the statistical properties of fairness metrics -- their virtues and limitations. Within this article, we attempt to contribute to the understanding of how ABROCA distributions and values are impacted by varying data conditions. This research helps refine fairness evaluation in learning analytics by helping to expand understanding of when research can confidently conclude when its statistical model is biased against specific demographic groups. While we conduct this analysis in the context of ABROCA, given its extensive application in learning analytics research, our simulation procedures can be applied to other bias metrics.

\section{Methods}

\subsection{Problem Definition and the ABROCA Metric}

A common metric for describing the predictive performance of a binary classifier is the area under the ROC Curve (AUC) metric \cite{bowers2019receiver}. The AUC metric has several desirable properties: a straightforward interpretation (i.e., the probability of the classifier correctly distinguishing a positive and negative case from each other) and describing performance adequately under imbalanced outcome classes \cite{jeni2013facing}. AUC ROC's robustness to imbalanced classes comes from weighting the true positive rate and false positive rate across all possible threshold values $t e [0, 1]$ to obtain a binary prediction. The Receiver Operating Characteristic curve (ROC) describes the relationship between the true positive rate and false different rate given different thresholds, with the AUC being the area under the ROC curve. The larger the AUC, the better the classifier's performance, with 0.5 corresponding to at-chance and 1 corresponding to perfect prediction. 

Consider a classifier with different predictive performance across two groups. For example, a dropout prediction model might exhibit less accurate performance for low SES students. How do we know if a classifier performs equally well between two groups? While the most straightforward (and common) method to compare performance would be by independently computing AUC (or another preferred performance metric) for both groups and comparing them, a simple difference between both AUCs has a key disadvantage: it does not capture performance differences at different thresholds. As Gardner et al. \cite{gardner2019evaluating} argue, if a classifier is more accurate for Group A at threshold $t_1$ but better for Group B at threshold $t_2$ (i.e., the ROC curves cross), overall AUC differences will cancel out these threshold-specific differences. Hence, Gardner et al. propose the Absolute Between-ROC Area (ABROCA) metric to describe classifier performances. 

Unlike the comparison of two AUCs, for which statistical tests and distribution of the test statistic (e.g., the difference in AUCs) is known \cite{demler2012misuse}, the ABROCA metric's statistical distribution and properties are, to the best of our knowledge, largely unknown. Like all statistical measures, the ABROCA metric is subject to fluctuations stemming from sampling, meaning that computing the ABROCA metric in a given sample will not always result in the same ABROCA estimation as another sample drawn from the same population. However, how much, and in what way can the ABROCA metric vary? To answer these questions, we simulate ABROCA under different conditions, as described next.

\subsection{Simulating and Sampling ABROCA}

First, we simulate data from two distribution pairs $(X_1, y_1)$, $(X_2, y_2)$ which describe an association equivalent to effect sizes $AUC_1$ and $AUC_2$. To do so, we sample from two normal distributions with a mean effect size of Cohen's $d$ (i.e., $M=0$, $SD=1$ and $M=d$, $SD=1$) with $d$ corresponding to a given AUC. In doing this, we follow transformations by Salgado \cite{salgado2018transforming} to convert AUCs into Cohen's $d$. The intuition behind this transformation is that all positive cases $y+$ will correspond to samples drawn from the separated normal distribution with mean $d$. In contrast, all negative cases $y-$ will correspond to the other distribution. The larger $d$, the easier the separation of $y+$ and $y-$ based on $X$ due to a larger separation in the average $X$. For simplicity, we consider $X$ to be a univariate distribution. While we note that many machine learning models use multivariate feature sets for prediction, ABROCA is only dependent on the correctness of the model prediction, not the properties of the predictors themselves, which is why we can simplify our sampling procedure for $X$. Importantly, this approach assumes that the AUC values fully reflect any differences between the ROC curves. In other words, any variations we see in the data are considered to be just due to random chance.

Following standard practices, we train a model on 80\% of the training data, splitting a joined data set of $(X_1, y_1)$, $(X_2, y_2)$ randomly. To keep our modeling procedure as simple (and fast to compute) as possible, we choose a logistic regression model with a univariate predictor of the outcome and do not consider hierarchical effects. The ABROCA metric is then evaluated by generalizing the fitted regression model to the 20\% holdout test set and computing an ABROCA metric based on the group assignment of each observation in the test set. In computing the ABROCA metric, we follow procedures aligning with open-source code in Gardner et al. \cite{gardner2019evaluating}. A summary of the algorithm is given below:

\begin{algorithm}[H]
\caption{Sampling ABROCA with Input: $AUC_1$, $AUC_2$, $N_{\text{total\_sample}}$ and Output: ABROCA metric}
\begin{enumerate}
    \item Transform $AUC_1$ and $AUC_2$ into Cohen's $d$ using the transformation from Salgado et al. \cite{salgado2018transforming}
    \item Sample $(X_1, y_1)$ from a standard normal distribution $(M=0, SD=1)$
    \item Sample $(X_2, y_2)$ from a normal distribution with $(M=d, SD=1)$, where $d$ corresponds to $AUC_2$
    \item Generate a training set of size $N_{\text{train}} = 0.8 \times N_{\text{total\_sample}}$
    \item Generate a test set of size $N_{\text{test}} = 0.2 \times N_{\text{total\_sample}}$
    \item Randomly split $(X_1, y_1)$ and $(X_2, y_2)$ into training $(X_{\text{train}}, y_{\text{train}})$ and test sets $(X_{\text{test}}, y_{\text{test}})$
    \item Fit a logistic regression model on $(X_{\text{train}}, y_{\text{train}})$
    \item Predict outcomes on $(X_{\text{test}}, y_{\text{test}})$ using the fitted logistic regression model
    \item Compute the ABROCA metric based on group assignment of test set observations \cite{gardner2019evaluating}
\end{enumerate}
\end{algorithm}

We acknowledge that various preprocessing and modeling techniques are often used to deal with imbalanced classes or groups, such as stratification and minority class oversampling. However, we do not employ these for simplicity and because their effect on predictive accuracy is not the focus of the present study. In this current work, we aim to capture ABROCA distributions for different data settings if a researcher were to train a standard machine learning model and leave the question of how preprocessing techniques impact these distributions for future work. %

\subsection{Experiments and Simulation Parameters}

Following Section 3.2, we program simulations for computing ABROCA values for different sampling specifications. We vary (a) the total sample size with 20\% representing the test set sample size on which ABROCA is computed, (b) the \% ratio of the minority population, (c) the \% ratio of the positive case in the outcome distribution, and (d) the AUC value of the minority condition. We consider the majority AUC to always equal 0.8, a level of performance generally considered acceptable for a binary classifier in education \cite{baker2024big}. For each parameter setting, we replicate each simulation 1,000 times to obtain a distribution of ABROCA values. By taking quantiles at 2.5\% and 97.5\% we establish 95\% confidence intervals in each ABROCA estimate, with the 50\% percentile representing a point estimation robust to skewed distributions.

We conducted three simulation experiments. For RQ1, we explored the distribution of ABROCA under no difference in AUCs between two groups by varying the total sample size from 500 to 9,500 in steps of 500 for finer granularity at lower ranges and from 10,000 to 100,000 in steps of 5,000. The minority population ratio and positive outcome ratio were both fixed at 50\%, with both groups sampled from AUCs of 0.8. For RQ2, we examined how ABROCA distributions change as differences in AUC ROC increase, replicating the setup of RQ1 but varying the minority group's AUC to 0.79, 0.75, 0.7, 0.6, and 0.5. For RQ3, we investigated how ABROCA distributions vary by minority class and outcome class imbalance. We varied the ratio of the minority population and the ratio of positive outcomes independently at values of 90\% and 50\%, using a fixed total sample size of 3,000 and minority group AUCs of 0.8 and 0.6.

\section{Results}

\subsection{RQ1: ABROCA Simulation Under No AUC Difference}

\begin{figure}[htpb]
    \centering
    \includegraphics[width=.805\linewidth]{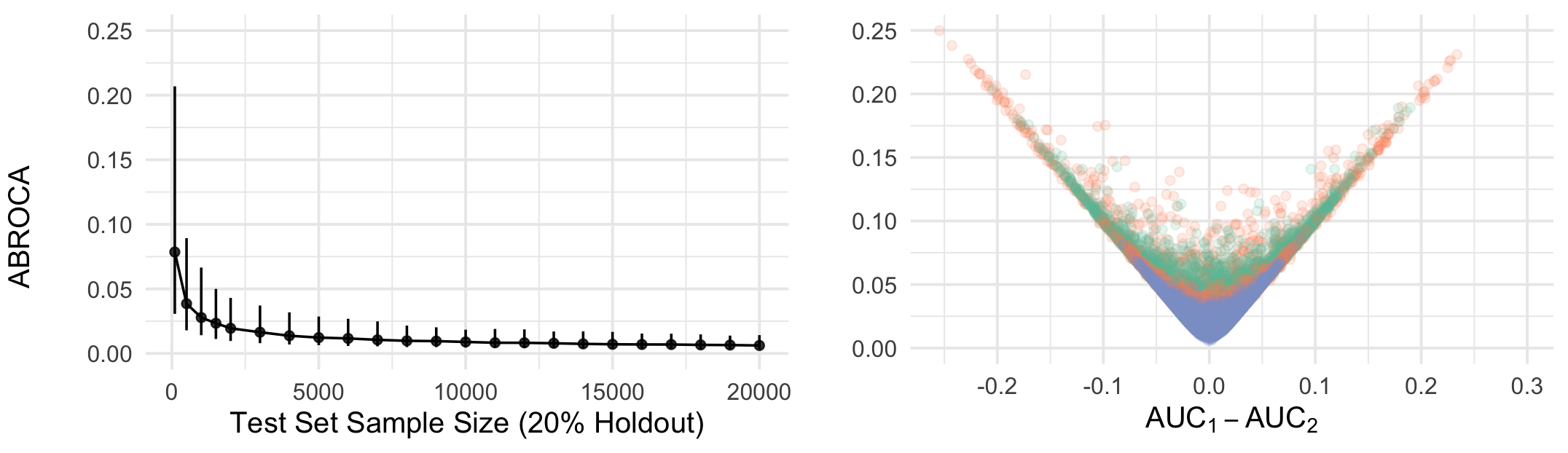}
    \caption{ABROCA distributions under no AUC difference for different test set sample sizes, including median point estimates and 95\% confidence intervals based on repeated simulations (left) and in relationship to AUC1-AUC2, for small (500-3,500; blue), medium (3500-6,000; orange), and large (6,000-9,500; green) test set sample sizes (right).}
    \label{fig:rq1}
\end{figure}

RQ1 focuses on the distribution of ABROCA under no population AUC differences. Three observations can be made based on Figure \ref{fig:rq1} (left). First, smaller sample sizes for the test set lead to larger distributions for ABROCA. While a test set size of 100 resulted in a median ABROCA of 0.08, a sample size of 500 corresponded to a median ABROCA of 0.04, and a sample size of 1,000 samples corresponded to a median ABROCA of 0.03. In other words, ABROCA approaches 0 with larger sample sizes. Second, the ABROCA statistic is skewed toward higher values for smaller sample sizes, meaning that the upper end of the distribution has a wider range than the lower end. Based on the mean minus the median of the distribution as a standard skewness measure, a test set sample size of 100 resulted in a skewness of 0.01, 500 in skewness of 0.004, 1,000 in 0.003, and so forth, approaching 0 with larger sample sizes. Third, the parametric confidence interval of the ABROCA distribution becomes smaller with larger sample sizes, which is expected as larger samples generally allow for more precise estimates. Notably, due to the positive skew of the ABROCA distribution at smaller sample sizes, some confidence intervals do not encompass point estimates obtained at larger sample sizes. For example, the lower bound of the confidence interval at 100 test set data points is 0.03. In that case, a researcher might conclude that the population ABROCA is larger than 0.03 with 95\% confidence, even when there is no difference in the population-level ROC curves. This issue is exemplified in Figure 1 (right): Even when there is no difference in AUCs or population-level curves, the ABROCA statistic can still yield relatively high values by chance. This is a property of ABROCA being strictly larger than $AUC_1$-$AUC_2$: random variation can only increase ABROCA, not decrease it.

\subsection{RQ2: ABROCA Simulation Under Different AUC Differences}

\begin{figure}[htpb]
    \centering
    \includegraphics[width=.775\linewidth]{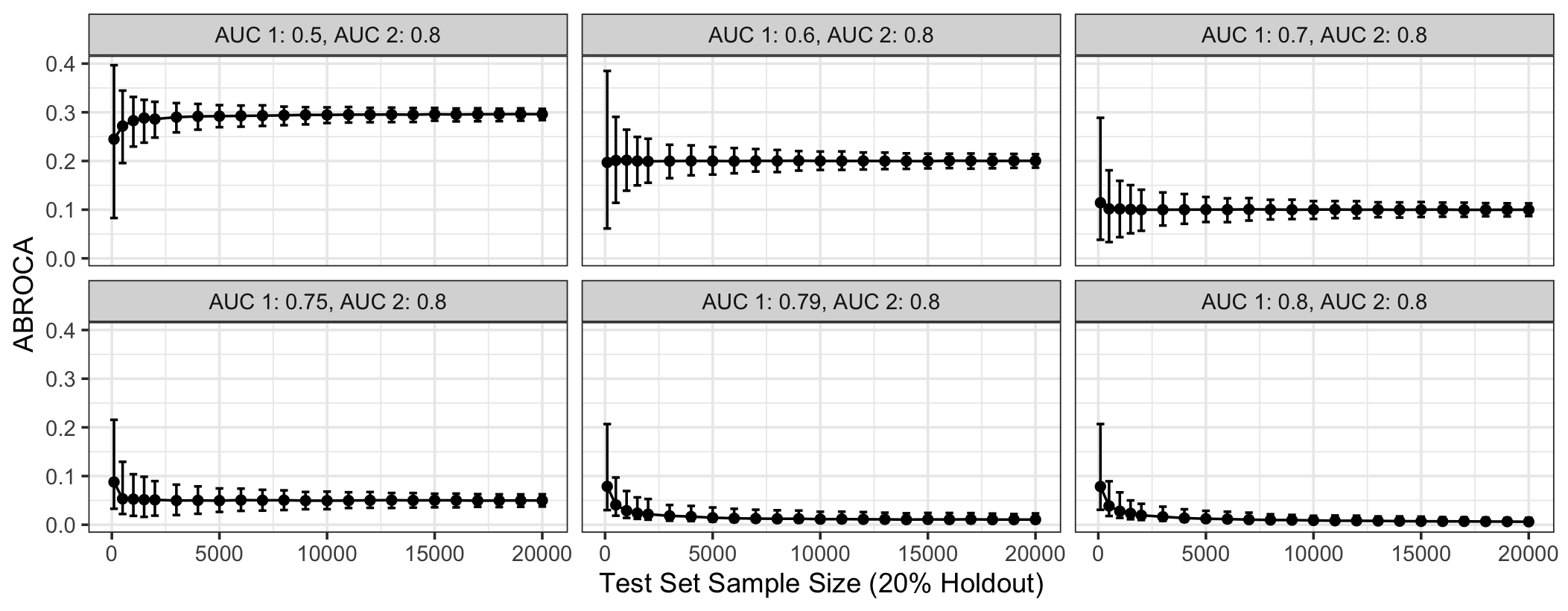}
    \caption{Distribution of the ABROCA statistic under different true AUC differences for different test set sample sizes, including median point estimates and 95\% confidence intervals based on repeated simulations.}
    \label{fig:rq2}
\end{figure}

RQ2 focuses on the distribution of ABROCA values under different true differences in AUCs. Based on Figure \ref{fig:rq2}, first, ABROCA values converge to the true difference in AUCs, as indicated by median estimates across increasing sample sizes. For example, ABROCA converges to 0.2 for a true difference in AUCs of 0.2 (middle top panel). Second, a skew in distributions can be observed at smaller sample sizes (similar to RQ1). This skew is predicated on the true AUC difference (when assuming in our simulation that there is no actual difference between population-level ROC curves beyond differences captured in overall AUC). Specifically, for a true AUC difference of larger than 0.3 (top left panel), ABROCA's distribution skew is negative. In all other cases, small sample sizes lead to positive skew. One exception is the AUC difference of 0.2 (between 0.6 and 0.8; top middle panel), with no or minimal skew. The skew was higher for smaller absolute differences in population AUCs (i.e., largest for equal AUCs, bottom right). Third, the rate at which the estimation of ABROCA becomes more accurate with more samples, as indicated by the size of the confidence intervals, does not seem to depend on the true AUC difference. In other words, the confidence interval sizes are approximately equally large across true AUC differences, given a sample size.

\begin{figure}[htpb]
    \centering
    \includegraphics[width=.725\linewidth]{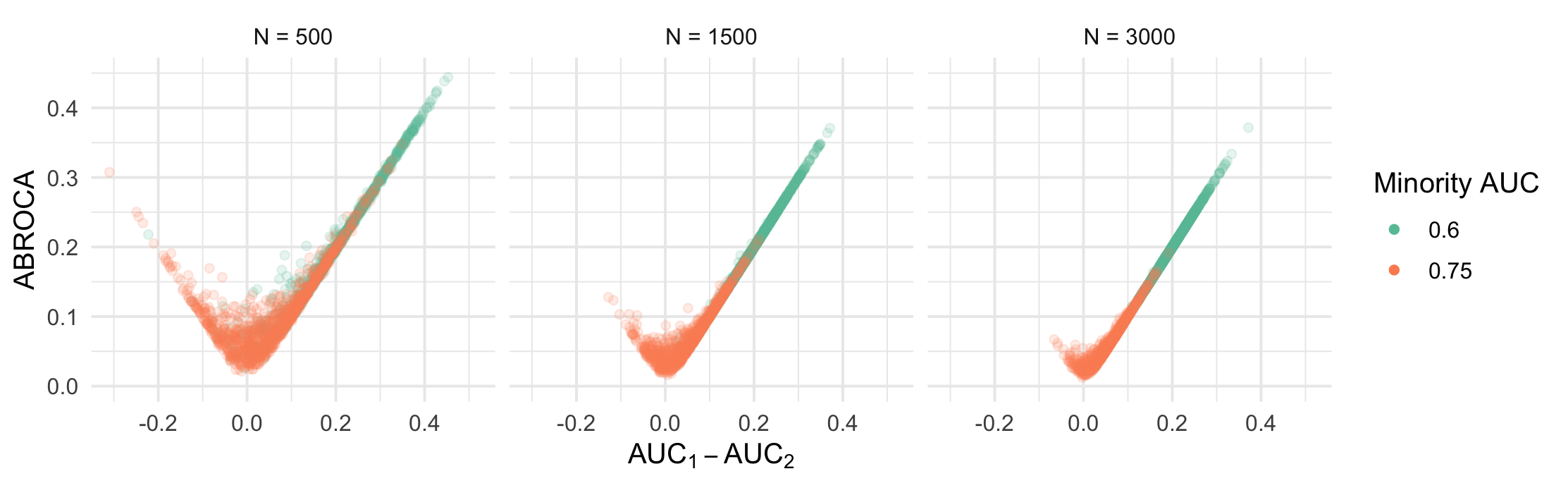}
    \caption{Sampled ABROCA values across three sample sizes and two population AUC differences for demonstration.}
    \label{fig:rq2-3}
\end{figure}

Figure \ref{fig:rq2-3} confirms the intuition that the closer two AUCs come to one another, the likelier they randomly cross by chance, skewing ABROCA values up. For more different AUC values, ABROCA converges toward $AUC_1$-$AUC_2$.

In summary, findings from RQ2 indicate that ABROCA approximates the true AUC difference for larger samples, while its skew depends on the proximity of the AUC values associated with each group.

\subsection{RQ3: ABROCA Simulations Under Different Majority Group and Class Ratios}

RQ3 focuses on how the distribution of ABROCA varies by the imbalance of outcome classes and population sizes. Based on Figure \ref{fig:rq3} below, the skewness of ABROCA identified in RQ1 is larger when the minority outcome class or the minority group is imbalanced (top row; compare leftmost column to other columns). Further, ABROCA was especially high by chance (and skewed) when both the minority class and outcome class were imbalanced (top row; rightmost column). In other words, the ABROCA statistic can yield especially high values due to chance for imbalanced data when the data is drawn from populations with equivalent ROC curves. We also find that balanced samples correspond to more precise estimates of the ABROCA metric, as indicated by less variation in the ABROCA estimate across samples (bottom row; leftmost column). In contrast, as the imbalance increases, the variation in ABROCA grows, suggesting less reliable estimates. This is particularly evident in cases where both the outcome and population sizes are imbalanced (bottom row, rightmost column), where the distribution flattens, implying higher uncertainty and variability.

\begin{figure}[htpb]
    \centering
    \includegraphics[width=.66\linewidth]{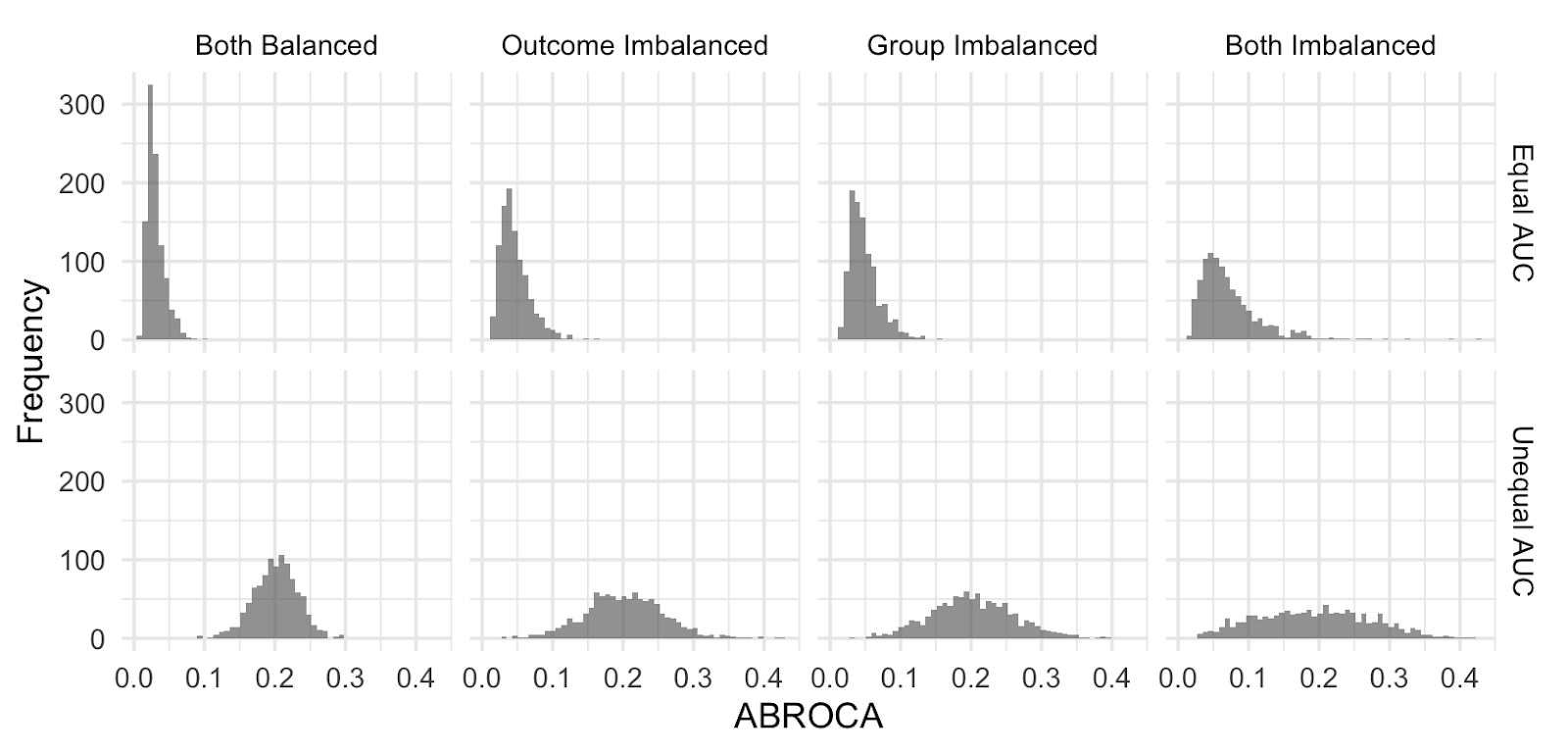}
    \caption{Histograms of sampled ABROCA values under balanced (50\%) and imbalanced (90\%) outcome and minority group classes as well as equal (0.8) and lower (0.6 vs. 0.8) AUC values in the minority group for 5,000 observations.}
    \label{fig:rq3}
\end{figure}

\section{Discussion}

ABROCA is a commonly used metric in learning analytics to describe classifier performance differences between groups. Unlike AUC ROC, it captures nuanced differences in classifier performance across decision-making thresholds, even if overall group-level performance is equal \cite{gardner2019evaluating}. However, ABROCA's distributional properties, especially for determining whether a classifier is overall biased, have not been studied, although ABROCA is often used to draw that type of conclusion. In this paper, we conducted simulations offering insights into how ABROCA behaves under various conditions applicable to common prediction tasks in the field, focusing on ABROCA differences based on chance.

\subsection{ABROCA is Skewed Toward Larger Values for More Similar AUCs and Smaller Sample Sizes (RQ1)}

Our findings show that ABROCA can be unexpectedly large, even when ROC curves—whose region differences increase ABROCA—are generated from the same population with no algorithmic bias. This seems to especially be a concern when overall differences in AUC and sample sizes are small (RQ1). In other words, when two AUCs are close to each other, they are more likely to cross over by chance. This is crucial as past applications of ABROCA in learning analytics did not consider this possible risk when using ABROCA to assess model bias (e.g., \cite{deho2024past,xu2024contexts}). Hence, to determine whether an obtained ABROCA constitutes a substantial difference in model performance between subgroups, these random overlaps need to be taken into account and perhaps adjusted for. Our open-source simulation code can help assess what level of ABROCA might be expected by chance for a specific case and eventually simulate at what threshold such an adjustment should be made to be able to reliably conclude an obtained ABROCA metric constitutes substantial bias. However, to evaluate ABROCA beyond its common use as a measure of overall model bias, it may be useful for future research to study individual crossover points where ROC curves intersect in more depth. Specifically, while our simulation provides insight into the average size of crossover regions that occur by chance, helping to establish a threshold for overall bias, it may be useful for future research to assess the reliability of the location of crossover regions. This aspect is a unique strength of ABROCA and merits further investigation.

\subsection{ABROCA Converges to $AUC_1 - AUC_2$ for Dissimilar AUCs: Its Most Useful Regions are Skewed (RQ2)}

As the sample size increases and AUC values become more different, ABROCA estimates converge toward the difference between the AUCs of the subpopulations (RQ2). This is notable because capturing bias that is not visible through the lens of overall performance differences is the key proposed benefit of ABROCA compared to AUC \cite{gardner2019evaluating}. Taken together with findings from RQ1, this introduces a conundrum: ABROCA is particularly valuable in detecting nuanced differences related to ROC crossover points that the simple difference of $AUC_1$-$AUC_2$ might miss. However, these very scenarios with crossover points are skewed and harder to detect accurately, requiring larger sample sizes and careful interpretation to avoid biased or skewed assessment of the degree of algorithmic bias.

\subsection{ABROCA Skew is Particularly Large When Group and Outcome Classes are Imbalanced (RQ3)}

In answering RQ3, we found that imbalanced outcome classes and minority groups make ABROCA more skewed (for more similar group-level AUCs). While our results also suggest that this issue is generally less serious for more balanced sample sizes, or larger minority class sizes, much recent research on algorithmic bias does not work with group sizes that our simulations deem sufficiently large (e.g., over 1,000 cases in the test set assuming balanced groups). Intersectional subsamples, for example, grouping students based on attribute combinations of race with gender \cite{zambrano2024investigating}, and minority student populations in general \cite{zambrano2024investigating} more commonly range in the hundreds rather than thousands.

Given that class imbalances in groups and outcomes are common in learning analytics (e.g., in outcome prediction), our results bear notable implications for the field. Firstly, there is a risk that studies using ABROCA may report false positive findings regarding performance differences and unfairness between groups (although how often false positive findings occur is not directly the subject of this study). The presence of crossover points in ROC curves, which might be due to noise rather than genuine performance disparities, could, in many cases, be due to random chance. By improving our estimation methods, we can more accurately assess and address genuine instances of algorithmic bias. A key area for future research will be to develop methods to establish statistical significance and reliability of algorithmic bias established by differences in ABROCA, incorporating thresholds established through simulations like those featured in this study or closed-form methods for determining critical quantiles in the ABROCA test statistic distribution.

\subsection{Limitations and Future Work}

There are several limitations to our simulation procedure. First, our simulation procedure assumed a single AUC per subgroup and normally distributed variables following recommendations in Salgado \cite{salgado2018transforming}. However, data sets often follow a range of distributions and exhibit other issues, such as missing data and selection effects. Hence, our findings may not match what is seen for ABROCA under more complex conditions. In future research, simulating data from a broader range of distributions and conditions could yield more accurate and generalizable insights into ABROCA's behavior in realistic scenarios. Future research could leverage existing models and datasets to perform simulations that better reflect real-world conditions using our open-source code. These efforts may include selecting input parameters of a simulation based on real-world data or bootstrapping data from their distributions. Second, our study focuses on the statistical properties of ABROCA without addressing the practical implications of observed biases. A statistically significant difference in classifier performance (or the lack thereof) may not translate into meaningful differences in practice. It is poorly understood how the type of differences that ABROCA captures—different biases in different model ranges—impact practice. Future work could simulate model downstream impact of performance differences on student outcomes (i.e., disparate impact), bridging the gap between statistical significance and practical relevance.

\section{Conclusion}

This study investigated the distributional properties of the ABROCA metric for assessing algorithmic bias. While ABROCA captures nuanced differences in classifier performance across thresholds, we observed some factors suggesting the use of caution in its interpretation as a primary measure of bias. Notably, ABROCA is skewed toward larger values when subgroup AUCs are similar and sample sizes are small, potentially leading to inflated bias estimates. As sample sizes increase and AUC differences widen, ABROCA converges toward the difference between subgroup AUCs (assuming no underlying regional differences between models), as crossover points by chance in these settings are rare. Outcome class and group imbalances exacerbate skewness and statistical uncertainty in ABROCA estimation. Our findings highlight the need for careful interpretation of ABROCA when used to assess overall algorithmic fairness.

\bibliographystyle{ACM-Reference-Format}
\bibliography{main}


\begin{thebibliography}{18}


\ifx \showCODEN    \undefined \def \showCODEN     #1{\unskip}     \fi
\ifx \showDOI      \undefined \def \showDOI       #1{#1}\fi
\ifx \showISBNx    \undefined \def \showISBNx     #1{\unskip}     \fi
\ifx \showISBNxiii \undefined \def \showISBNxiii  #1{\unskip}     \fi
\ifx \showISSN     \undefined \def \showISSN      #1{\unskip}     \fi
\ifx \showLCCN     \undefined \def \showLCCN      #1{\unskip}     \fi
\ifx \shownote     \undefined \def \shownote      #1{#1}          \fi
\ifx \showarticletitle \undefined \def \showarticletitle #1{#1}   \fi
\ifx \showURL      \undefined \def \showURL       {\relax}        \fi
\providecommand\bibfield[2]{#2}
\providecommand\bibinfo[2]{#2}
\providecommand\natexlab[1]{#1}
\providecommand\showeprint[2][]{arXiv:#2}

\bibitem[Baker(2024)]%
        {baker2024big}
\bibfield{author}{\bibinfo{person}{Ryan~S. Baker}.} \bibinfo{year}{2024}\natexlab{}.
\newblock \bibinfo{booktitle}{\emph{Big Data and Education} (\bibinfo{edition}{8th} ed.)}.
\newblock \bibinfo{publisher}{University of Pennsylvania}, \bibinfo{address}{Philadelphia, PA}.
\newblock


\bibitem[Baker and Hawn(2022)]%
        {baker2022algorithmic}
\bibfield{author}{\bibinfo{person}{Ryan~S. Baker} {and} \bibinfo{person}{Aaron Hawn}.} \bibinfo{year}{2022}\natexlab{}.
\newblock \showarticletitle{Algorithmic Bias in Education}.
\newblock \bibinfo{journal}{\emph{International Journal of Artificial Intelligence in Education}} (\bibinfo{year}{2022}), \bibinfo{pages}{1--41}.
\newblock


\bibitem[Barocas et~al\mbox{.}(2020)]%
        {barocas2020hidden}
\bibfield{author}{\bibinfo{person}{Solon Barocas}, \bibinfo{person}{Andrew~D Selbst}, {and} \bibinfo{person}{Manish Raghavan}.} \bibinfo{year}{2020}\natexlab{}.
\newblock \showarticletitle{The hidden assumptions behind counterfactual explanations and principal reasons}. In \bibinfo{booktitle}{\emph{Proceedings of the 2020 Conference on Fairness, Accountability, and Transparency}}. \bibinfo{pages}{80--89}.
\newblock


\bibitem[Bowers and Zhou(2019)]%
        {bowers2019receiver}
\bibfield{author}{\bibinfo{person}{Alex~J Bowers} {and} \bibinfo{person}{Xiaoliang Zhou}.} \bibinfo{year}{2019}\natexlab{}.
\newblock \showarticletitle{Receiver Operating Characteristic (ROC) Area Under the Curve (AUC): A Diagnostic Measure for Evaluating the Accuracy of Predictors of Education Outcomes}.
\newblock \bibinfo{journal}{\emph{Journal of Education for Students Placed at Risk (JESPAR)}} \bibinfo{volume}{24}, \bibinfo{number}{1} (\bibinfo{year}{2019}), \bibinfo{pages}{20--46}.
\newblock


\bibitem[Deho et~al\mbox{.}(2024)]%
        {deho2024past}
\bibfield{author}{\bibinfo{person}{Oscar~Blessed Deho}, \bibinfo{person}{Lin Liu}, \bibinfo{person}{Jiuyong Li}, \bibinfo{person}{Jixue Liu}, \bibinfo{person}{Chen Zhan}, {and} \bibinfo{person}{Srecko Joksimovic}.} \bibinfo{year}{2024}\natexlab{}.
\newblock \showarticletitle{When the past!= the future: Assessing the Impact of Dataset Drift on the Fairness of Learning Analytics Models}.
\newblock \bibinfo{journal}{\emph{IEEE Transactions on Learning Technologies}} (\bibinfo{year}{2024}).
\newblock


\bibitem[Demler et~al\mbox{.}(2012)]%
        {demler2012misuse}
\bibfield{author}{\bibinfo{person}{Olga~V Demler}, \bibinfo{person}{Michael~J Pencina}, {and} \bibinfo{person}{Ralph~B D'Agostino~Sr}.} \bibinfo{year}{2012}\natexlab{}.
\newblock \showarticletitle{Misuse of DeLong Test to Compare AUCs for Nested Models}.
\newblock \bibinfo{journal}{\emph{Statistics in Medicine}} \bibinfo{volume}{31}, \bibinfo{number}{23} (\bibinfo{year}{2012}), \bibinfo{pages}{2577--2587}.
\newblock


\bibitem[Friedman and Nissenbaum(1996)]%
        {friedman1996bias}
\bibfield{author}{\bibinfo{person}{Batya Friedman} {and} \bibinfo{person}{Helen Nissenbaum}.} \bibinfo{year}{1996}\natexlab{}.
\newblock \showarticletitle{Bias in Computer Systems}.
\newblock \bibinfo{journal}{\emph{ACM Transactions on Information Systems (TOIS)}} \bibinfo{volume}{14}, \bibinfo{number}{3} (\bibinfo{year}{1996}), \bibinfo{pages}{330--347}.
\newblock


\bibitem[Gardner et~al\mbox{.}(2019)]%
        {gardner2019evaluating}
\bibfield{author}{\bibinfo{person}{Josh Gardner}, \bibinfo{person}{Christopher Brooks}, {and} \bibinfo{person}{Ryan Baker}.} \bibinfo{year}{2019}\natexlab{}.
\newblock \showarticletitle{Evaluating the Fairness of Predictive Student Models Through Slicing Analysis}. In \bibinfo{booktitle}{\emph{Proceedings of the 9th International Conference on Learning Analytics \& Knowledge}}. \bibinfo{pages}{225--234}.
\newblock


\bibitem[Jeni et~al\mbox{.}(2013)]%
        {jeni2013facing}
\bibfield{author}{\bibinfo{person}{L{\'a}szl{\'o}~A Jeni}, \bibinfo{person}{Jeffrey~F Cohn}, {and} \bibinfo{person}{Fernando De~La~Torre}.} \bibinfo{year}{2013}\natexlab{}.
\newblock \showarticletitle{Facing Imbalanced Data: Recommendations for the Use of Performance Metrics}. In \bibinfo{booktitle}{\emph{2013 Humaine Association Conference on Affective Computing and Intelligent Interaction}}. IEEE, \bibinfo{pages}{245--251}.
\newblock


\bibitem[Jiang and Pardos(2021)]%
        {jiang2021towards}
\bibfield{author}{\bibinfo{person}{Weijie Jiang} {and} \bibinfo{person}{Zachary~A Pardos}.} \bibinfo{year}{2021}\natexlab{}.
\newblock \showarticletitle{Towards Equity and Algorithmic Fairness in Student Grade Prediction}. In \bibinfo{booktitle}{\emph{Proceedings of the 2021 AAAI/ACM Conference on AI, Ethics, and Society}}. \bibinfo{pages}{608--617}.
\newblock


\bibitem[Kizilcec and Lee(2022)]%
        {kizilcec2022algorithmic}
\bibfield{author}{\bibinfo{person}{Ren{\'e}~F Kizilcec} {and} \bibinfo{person}{Hansol Lee}.} \bibinfo{year}{2022}\natexlab{}.
\newblock \showarticletitle{Algorithmic Fairness in Education}.
\newblock In \bibinfo{booktitle}{\emph{The Ethics of Artificial Intelligence in Education}}. \bibinfo{publisher}{Routledge}, \bibinfo{pages}{174--202}.
\newblock


\bibitem[Salgado(2018)]%
        {salgado2018transforming}
\bibfield{author}{\bibinfo{person}{Jes{\'u}s~F Salgado}.} \bibinfo{year}{2018}\natexlab{}.
\newblock \showarticletitle{Transforming the Area under the Normal Curve (AUC) into Cohen’s d, Pearson’s r pb , Odds-Ratio, and Natural Log Odds-Ratio: Two Conversion Tables}.
\newblock \bibinfo{journal}{\emph{European Journal of Psychology Applied to Legal Context}} \bibinfo{volume}{10}, \bibinfo{number}{1} (\bibinfo{year}{2018}), \bibinfo{pages}{35--47}.
\newblock


\bibitem[Sha et~al\mbox{.}(2021)]%
        {sha2021assessing}
\bibfield{author}{\bibinfo{person}{Lele Sha}, \bibinfo{person}{Mladen Rakovic}, \bibinfo{person}{Alexander Whitelock-Wainwright}, \bibinfo{person}{David Carroll}, \bibinfo{person}{Victoria~M Yew}, \bibinfo{person}{Dragan Gasevic}, {and} \bibinfo{person}{Guanliang Chen}.} \bibinfo{year}{2021}\natexlab{}.
\newblock \showarticletitle{Assessing Algorithmic Fairness in Automatic Classifiers of Educational Forum Posts}. In \bibinfo{booktitle}{\emph{Artificial Intelligence in Education: 22nd International Conference, AIED 2021, Utrecht, The Netherlands, June 14--18, 2021, Proceedings, Part I}}. Springer, \bibinfo{pages}{381--394}.
\newblock


\bibitem[{\v{S}}v{\'a}bensk{\`y} et~al\mbox{.}(2024)]%
        {vsvabensky2024evaluating}
\bibfield{author}{\bibinfo{person}{Valdemar {\v{S}}v{\'a}bensk{\`y}}, \bibinfo{person}{M{\'e}lina Verger}, \bibinfo{person}{Maria Mercedes~T Rodrigo}, \bibinfo{person}{Clarence James~G Monterozo}, \bibinfo{person}{Ryan~S Baker}, \bibinfo{person}{Miguel Zenon Nicanor~Lerias Saavedra}, \bibinfo{person}{S{\'e}bastien Lall{\'e}}, {and} \bibinfo{person}{Atsushi Shimada}.} \bibinfo{year}{2024}\natexlab{}.
\newblock \showarticletitle{Evaluating Algorithmic Bias in Models for Predicting Academic Performance of Filipino Students}. In \bibinfo{booktitle}{\emph{17th International Conference on Educational Data Mining (EDM 2024)}}.
\newblock


\bibitem[Verger et~al\mbox{.}(2023)]%
        {verger2023your}
\bibfield{author}{\bibinfo{person}{M{\'e}lina Verger}, \bibinfo{person}{S{\'e}bastien Lall{\'e}}, \bibinfo{person}{Fran{\c{c}}ois Bouchet}, {and} \bibinfo{person}{Vanda Luengo}.} \bibinfo{year}{2023}\natexlab{}.
\newblock \showarticletitle{Is Your Model ``MADD''? A Novel Metric to Evaluate Algorithmic Fairness for Predictive Student Models}. In \bibinfo{booktitle}{\emph{16th International Conference on Educational Data Mining (EDM 2023)}}.
\newblock


\bibitem[Verma and Rubin(2018)]%
        {verma2018fairness}
\bibfield{author}{\bibinfo{person}{Sahil Verma} {and} \bibinfo{person}{Julia Rubin}.} \bibinfo{year}{2018}\natexlab{}.
\newblock \showarticletitle{Fairness Definitions Explained}. In \bibinfo{booktitle}{\emph{Proceedings of the International Workshop on Software Fairness}}. \bibinfo{pages}{1--7}.
\newblock


\bibitem[Xu et~al\mbox{.}(2024)]%
        {xu2024contexts}
\bibfield{author}{\bibinfo{person}{Zhen Xu}, \bibinfo{person}{Joseph Olson}, \bibinfo{person}{Nicole Pochinki}, \bibinfo{person}{Zhijian Zheng}, {and} \bibinfo{person}{Renzhe Yu}.} \bibinfo{year}{2024}\natexlab{}.
\newblock \showarticletitle{Contexts Matter but How? Course-Level Correlates of Performance and Fairness Shift in Predictive Model Transfer}. In \bibinfo{booktitle}{\emph{Proceedings of the 14th Learning Analytics and Knowledge Conference}}. \bibinfo{pages}{713--724}.
\newblock


\bibitem[Zambrano et~al\mbox{.}(2024)]%
        {zambrano2024investigating}
\bibfield{author}{\bibinfo{person}{Andres~Felipe Zambrano}, \bibinfo{person}{Jiayi Zhang}, {and} \bibinfo{person}{Ryan~S Baker}.} \bibinfo{year}{2024}\natexlab{}.
\newblock \showarticletitle{Investigating Algorithmic Bias on Bayesian Knowledge Tracing and Carelessness Detectors}. In \bibinfo{booktitle}{\emph{Proceedings of the 14th Learning Analytics and Knowledge Conference}}. \bibinfo{pages}{349--359}.
\newblock


\end{thebibliography}

\end{document}